\documentclass[10pt,twocolumn,letterpaper]{article}
\usepackage{multirow}

 \usepackage{iccv}              % To produce the CAMERA-READY version
\definecolor{iccvblue}{rgb}{0.21,0.49,0.74}
\usepackage[pagebackref,breaklinks,colorlinks,allcolors=iccvblue]{hyperref}

\def\paperID{2} % *** Enter the Paper ID here
\def\confName{ICCV}
\def\confYear{2025}

\title{AI-Derived Structural Building Intelligence for Urban Resilience: \\ An Application in Saint Vincent and the Grenadines}

\author{Isabelle Tingzon, Yoji Toriumi, Caroline Gevaert\\
The World Bank Group\\
{\tt\small \{tisabelle, ytoriumi, cgevaert\}@worldbank.org}
}
\begin{document}
\maketitle

\begin{abstract}
Detailed structural building information is used to estimate potential damage from hazard events like cyclones, floods, and landslides, making them critical for urban resilience planning and disaster risk reduction. However, such information is often unavailable in many small island developing states (SIDS) in climate-vulnerable regions like the Caribbean. To address this data gap, we present an AI-driven workflow to automatically infer rooftop attributes from high-resolution satellite imagery, with Saint Vincent and the Grenadines as our case study. Here, we compare the utility of geospatial foundation models combined with shallow classifiers against fine-tuned deep learning models for rooftop classification. Furthermore, we assess the impact of incorporating additional training data from neighboring SIDS to improve model performance. Our best models achieve F1 scores of 0.88 and 0.83 for roof pitch and roof material classification, respectively. Combined with local capacity building, our work aims to provide SIDS with novel capabilities to harness AI and Earth Observation (EO) data to enable more efficient, evidence-based urban governance.

\end{abstract}

\section{Introduction}
Comprehensive information on structural building attributes is critical for effective urban resilience planning, targeted interventions, and strategic investment decisions. However, such detailed data are often lacking in low- and middle-income countries (LMICs), particularly in small island developing states (SIDS), due to the high costs associated with carrying out large-scale building surveys. For Caribbean SIDS, which are highly exposed to hurricanes, earthquakes, landslides, and flooding, this data gap poses a major challenge to enforcing building regulatory codes and ensuring the resilience of critical infrastructure to natural hazards \cite{oecs2025}.

While prior research has made progress in addressing these challenges within the Caribbean context, most have relied on very high-resolution aerial imagery (2 to 10 cm/px), achieving F1 scores between 0.88 and 0.92 \cite{tingzon2024mapping,tingzon2023fusing}. This raises the question of whether similar performance can be achieved using relatively lower-resolution (30 to 60 cm/px) satellite imagery in countries lacking very high-resolution data. Furthermore, earlier works have not explored the use of modern geospatial foundation models \cite{reed2023scale,ayush2021geography}, which have the potential to accelerate model development for rooftop classification in novel geographic contexts, provided that their performance is comparable to that of traditional fine-tuned deep learning approaches.

To address these challenges, we propose an end-to-end, AI-driven workflow for the automated extraction of structural building attributes from high-resolution Maxar satellite imagery, using Saint Vincent and the Grenadines as our case study. Our study compares the performance of geospatial foundation models and shallow classifiers with that of traditional fine-tuned deep learning models for classifying rooftop attributes. Furthermore, we evaluate the impact of incorporating additional training data from neighboring countries, namely Saint Lucia and Dominica, on model performance. Finally, leveraging our best-performing models, we generate a structural baseline inventory by deploying our roof classification models across over 40K building footprints within in Saint Vincent and the Grenadines. We publicly release the first-ever building classification map for Saint Vincent and the Grenadines\footnote{\url{https://datacatalog.worldbank.org/search/dataset/0065199/Rooftop-classification-for-OECS-countries}}.

Through strong stakeholder engagement and local capacity building, our work aims to equip SIDS with a novel capability to harness AI and Earth Observation (EO) to assess building vulnerability, monitor regulatory compliance, and support resilient asset management. For city governments, this approach represents a transformative tool for data-driven planning and disaster risk management,  allowing for scalable assessments and offering a path toward more proactive and efficient urban governance.

\begin{figure*}
    \centering
    \includegraphics[width=1\linewidth]{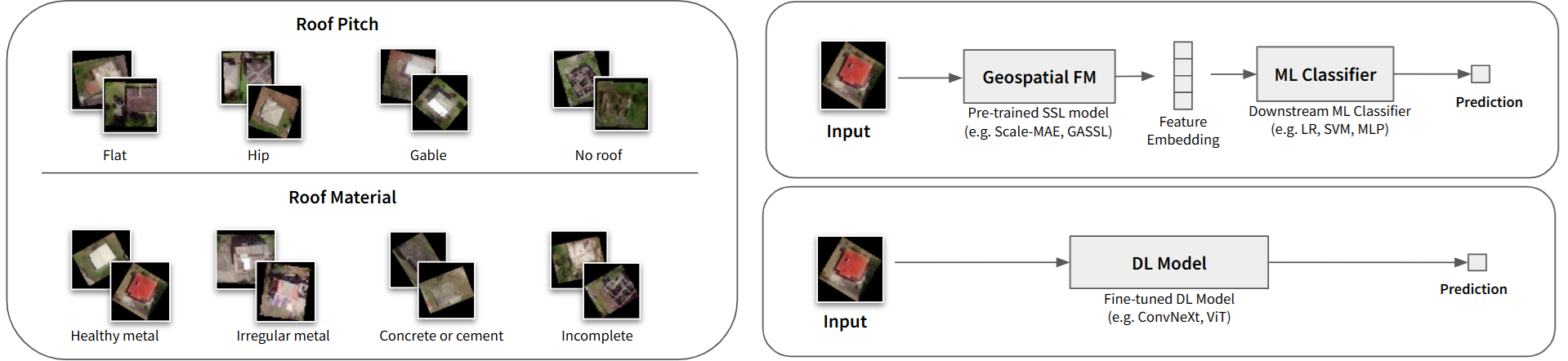}
    \caption{An overview of the roof classes in Saint Vincent and the Grenadines (left) and model experimentation (right) using geospatial foundation models (FM) combined with machine learning (ML) classifiers (top) and fine-tuned deep learning (DL) models (bottom).}
    \label{fig:enter-label}
\end{figure*}

\section{Data}
We begin by leveraging publicly available geospatial data sources for Saint Vincent and the Grenadines, namely (1) high-resolution Maxar satellite images obtained from OpenAerialMap (OAM) \cite{smith2015openaerialmap} and (2) building footprints polygons from Microsoft Building Footprints \cite{mbf2023}. The satellite images from OAM were captured at various time points between 2017 and 2021, with spatial resolutions ranging from 32 to 54 cm/px.  All satellite images were merged to create a single mosaicked composite with complete, nationwide coverage of Saint Vincent and the Grenadines. 

From the composite satellite image, we then cropped the minimum bounding rectangle of each building footprint, scaled by
a factor of 2. This scaling was applied to account for misalignments between the building footprints and the underlying satellite imagery, increasing the likelihood that the actual building would be captured within the cropped image. To identify buildings in areas obscured by high cloud cover, we used Canny edge detection to extract the edges of objects within the images \cite{canny1986computational}. Images with little to no edges detected were subsequently flagged and removed from our set of images for annotation.

To generate a diverse and representative ground truth dataset for Saint Vincent and the Grenadines, we began by randomly selecting 250 tiles of size 500 m $\times$ 500 m across the country. A group of three GIS experts were then tasked with annotating all buildings within a subset of selected tiles via visual interpretation of the RGB satellite images. Consistent with previous works \cite{tingzon2024mapping}, buildings were annotated based on two main rooftop characteristics: (1) roof material (healthy metal, irregular metal, concrete/cement, incomplete) and (2) roof pitch (hip, gable, flat, and no roof). 

In line with data-centric learning \cite{datacentric}, we increased the number of samples in the minority classes (e.g., irregular metal, incomplete) by leveraging feature embeddings from the pre-trained EO foundation model ScaleMAE \cite{reed2023scale} to identify the top-k images most similar to a given query image, based on cosine similarity. For Saint Vincent and the Grenadines, we manually reviewed the top 25 most similar images retrieved for selected query images and added the correctly matched samples to our dataset. As a result, our final dataset comprised 3,243 labeled buildings in Saint Vincent and the Grenadines, the class distributions of which are presented in Table 1. 

\begin{table}[]
\footnotesize
\caption{Class distribution of roof type and roof material across Saint Vincent and the Grenadines (\textbf{VCT}), Saint Lucia (\textbf{LCA}), and Dominica (\textbf{DCA}).}
\label{table:classdist}
\centering
\begin{tabular}{@{}p{0.5cm}p{1.845cm}rrrrr@{}}
\toprule
& & \textbf{VCT} & \textbf{LCA} & \textbf{DCA}  & \textbf{Total} \\ \midrule 
\multirow{4}*{\rotatebox{90}{\textbf{\footnotesize{{\begin{tabular}[c]{@{}c@{}}Roof \\ Pitch\end{tabular}}}}}}  
& Gable    & 1,717 & 2,347 & 2,172 &  6,236  \\   
& Hip   & 902 & 1,089 & 1,251 &  3,242 \\
& Flat     & 487 & 456 & 1,625 & 2,568 \\         
& No Roof & 137 & 269 & 1,190 &  1,596 \\
\midrule
\multirow{4}*{\rotatebox{90}{\textbf{\footnotesize{{\begin{tabular}[c]{@{}c@{}}Roof \\ Material\end{tabular}}}}}}  
& Healthy metal   & 2,372 & 2,396 & 1,934 & 6,702 \\
& Concrete/cement & 423 & 328 & 1,240 &  1,991  \\
& Irregular metal & 295 & 1,113 & 1,733 &  3,141 \\
& Incomplete    & 153 & 324 & 1,331 &  1,808 \\
\midrule
& \textbf{Total}   & 3,243 & 4,161 & 6,238 & 13,642  \\ \bottomrule
\end{tabular}
\end{table}

\subsection*{Data Split}
For model training and evaluation, we split the dataset into 80\% training and 20\% testing sets using a stratified group shuffle split, where buildings were grouped according to the 500 m $\times$ 500 m tile in which they belong. This approach preserves class distributions within each split while ensuring that all buildings within the same tile are assigned to the same split, thereby reducing the risk of data leakage.

\subsection*{Saint Lucia and Dominica Data}
To determine whether additional data from neighboring small island developing states would improve model performance for Saint Vincent and the Grenadines, we augment our training data with additional labeled aerial images from Saint Lucia and Dominica, as introduced in \citep{tingzon2024mapping}. These datasets are comprised of high-resolution aerial orthophotos within Saint Lucia and Dominica, with spatial resolutions of 10 cm/px and 20 cm/px, respectively. The datasets also include very high-resolution drone imagery with spatial resolutions ranging from 2 to 7 cm/px for selected areas within Saint Lucia and Dominica. 

For consistency with the spatial resolution of the aerial images in Saint Vincent and the Grenadines, we decreased the resolution of the 10 cm/px aerial orthophotos in Saint Lucia by a factor of 5, the 20 cm/px orthophotos in Dominica by a factor of 2.5, and all drone images by a factor of 6. Finally, we removed the blue tarpaulin roof material class from the Dominica dataset, as it was deemed relevant only in post-disaster contexts. The combined training datasets of Saint Lucia, Dominica, and Saint Vincent and the Grenadines comprised a total of 12,860 images. We detail the class distributions across the three datasets, both separately and combined, in Table 1.

\section{Methodology}
We start by evaluating the utility of geospatial foundation models as feature extractors for the downstream task of rooftop classification. Feature embeddings from pre-trained foundation models allow for accelerated model development using only shallow classifiers, making them valuable in resource-constrained settings. We then benchmark these results against that of the more traditional approach of fine-tuning deep learning models for image classification. 

\subsection*{Foundation models + shallow classifiers}
We leveraged the pre-trained weights of geospatial foundation models, namely Scale-MAE \cite{reed2023scale} and Geography-Aware Self-Supervised Learning (GASSL) \cite{ayush2021geography}, as provided through the TorchGeo library \cite{stewart2022torchgeo}. Both models were pre-trained in a self-supervised manner on the Functional Map of the World (FMoW) dataset \cite{christie2018functional}, which contains high-resolution satellite imagery from across the globe, with the goal of learning low-dimensional feature representations that capture contextual spatial information relevant for downstream remote sensing classification tasks.

For each image in our dataset, we extracted feature embeddings of size 1,024 using Scale-MAE and size 2,048 using GASSL. The feature embeddings were then used as input to shallow classifiers, including logistic regression (LR), support vector machines (SVM), and multilayer perceptrons (MLP) for the downstream task of rooftop classification. For each classifier, we implemented hyperparameter tuning on the training set using stratified group 5-fold cross-validation (CV).  For more information on the hyperparameter tuning, see Appendix \ref{appendix:a}.

\subsection*{Fine-tuning deep learning models}
We selected three variants of ConvNeXt (i.e., small, base, and large) \cite{liu2022convnet} and two variants of Vision Transformers (ViT) (i.e., base and large) \cite{dosovitskiy2020image} as our base architectures for deep learning model development. All models were pre-trained on the ImageNet dataset \cite{deng2009imagenet} and fine-tuned using multi-class cross-entropy loss. 

To prepare the data for model training, all input images were zero-padded to a square based on the maximum value between the width and height of the image, resized to 224 x 224 px, and normalized using the mean and standard deviation of the ImageNet dataset. Data augmentation was done in the form of random vertical and horizontal image flips and rotations ranging from $-90^{\circ}$ to $90^{\circ}$. For model training, we used the Adam optimizer, set the batch size to 16, and used an initial learning rate of 1e-5, which was reduced by a factor of 0.1 after every 7 epochs of no improvement. All models were trained for a maximum of 30 epochs with early stopping once the learning rate fell below 1e-7.

\section{Results and Discussion}
For model evaluation, we report the macro-averaged precision, recall, accuracy, and F1 score, with the latter as our primary metric of performance for model selection. Table \ref{table:results1} presents the test set results for each roof classification task, using models trained only on the Saint Vincent and the Grenadines training set.

\begin{table}[]
\small
\caption{Comparison of the test set results for (a) roof pitch classification and (b) roof material classification using models trained only on the training data for Saint Vincent and the Grenadines.}
\label{table:results1}
\resizebox{0.475\textwidth}{!}{
\begin{tabular}{p{2.5cm}ccccc@{}}
\multicolumn{4}{@{}l}{(a) Roof pitch classification}\\
\toprule
& \textbf{F1 score} & \textbf{Precision} & \textbf{Recall} &  \textbf{Accuracy} \\ \midrule 
GASSL+LR    & 0.748 & 0.776  & 0.726 & 0.781 \\
GASSL+SVM    & 0.723 & 0.727  & 0.720 & 0.757 \\
GASSL+MLP    & 0.701 & 0.698  & 0.706 & 0.744 \\
Scale-MAE+LR    & 0.594 & 0.632  & 0.568 & 0.692 \\
Scale-MAE+SVM    & 0.604 & 0.584  & 0.637 & 0.674 \\
Scale-MAE+MLP    & 0.617 & 0.624  & 0.618 & 0.679 \\
\midrule 
ConvNeXt-small    & 0.838 & 0.846  & 0.846  & 0.863  \\
ConvNeXt-base    & \textbf{0.858} & 0.866  & 0.859 & 0.880  \\              
ConvNeXt-large    & 0.856  & 0.879  & 0.838  & 0.871  \\
ViT-base    & 0.829 & 0.827  & 0.841 & 0.836 \\              
ViT-large    & 0.795 & 0.822  & 0.774 & 0.813 \\
\bottomrule
\end{tabular}}
\medskip \\
\resizebox{0.475\textwidth}{!}{
\begin{tabular}{p{2.5cm}ccccc@{}}
\multicolumn{4}{@{}l}{(b) Roof material classification}\\
\toprule
& \textbf{F1 score} & \textbf{Precision} & \textbf{Recall}  & \textbf{Accuracy} \\ \midrule
GASSL+LR    & 0.676 & 0.741  & 0.630 & 0.836 \\
GASSL+SVM    & 0.700 & 0.729  & 0.680 & 0.837 \\
GASSL+MLP    & 0.697 & 0.740  & 0.663 & 0.837 \\
Scale-MAE+LR    & 0.598 & 0.619  & 0.581 & 0.803 \\
Scale-MAE+SVM    & 0.560 & 0.588  & 0.617 & 0.772 \\
Scale-MAE+MLP    & 0.506 &  0.508 & 0.518 & 0.739 \\
\midrule
ConvNeXt-small    & 0.822 & 0.854  & 0.794 & 0.906 \\
ConvNeXt-base    & 0.828 & 0.839  & 0.820 & 0.907 \\              
ConvNeXt-large    & \textbf{0.835 }& 0.868  & 0.806 & 0.912  \\
ViT-base    &  0.818 & 0.856  & 0.787 & 0.904  \\              
ViT-large    & 0.777 & 0.817  & 0.744 & 0.884  \\
\bottomrule
\end{tabular}}
\end{table}

Our results suggest that although geospatial foundation models are useful for facilitating rapid model development, their performance for roof classification are still outperformed by that of traditionally fine-tuned deep learning models. Specifically, the best-performing foundation model + shallow classifier combinations achieve an F1 score of 0.748 for roof pitch classification (GASSL+LR) and 0.7 for roof material classification (GASSL+SVM). In comparison, the best fine-tuned deep learning models reach F1 scores of up to 0.858 for roof pitch classification and 0.835 for roof material classification. Therefore, we focus exclusively on fine-tuning deep learning models in subsequent experiments for model improvement.

Next, we examine whether incorporating additional training data from Saint Lucia and Dominica would improve model performance for Saint Vincent and the Grenadines. As shown in Table \ref{table:results2}, fine-tuning deep learning models on the combined training sets led to performance improvements for roof pitch classification, with the best F1 score increasing from 0.858 (using only local data) to 0.878 (using combined, regional data). However, for roof material classification, adding data from Saint Lucia and Dominica did not yield improvements, with the best F1 score slightly decreasing from 0.835 to 0.827.

\begin{table}[]
\small
\caption{Comparison of the test set results for (a) roof pitch classification and (b) roof material classification using deep learning models fine-tuned on the combined training datasets of Dominica, Saint Lucia, and Saint Vincent and the Grenadines.}
\label{table:results2}
\resizebox{0.475\textwidth}{!}{
\begin{tabular}{p{2.5cm}ccccc@{}}
\multicolumn{4}{@{}l}{(a) Roof pitch classification}\\
\toprule
& \textbf{F1 score} & \textbf{Precision} & \textbf{Recall} &  \textbf{Accuracy} \\ \midrule 
ConvNeXt-small    & 0.874 & 0.895  & 0.861 & 0.878 \\
ConvNeXt-base    & 0.868 & 0.880  & 0.859 & 0.872 \\              
ConvNeXt-large    & \textbf{0.878} & 0.892  & 0.867 & 0.885 \\
ViT-base    & 0.821 & 0.835  & 0.810 & 0.839 \\              
ViT-large    &  0.810 & 0.831  & 0.799 & 0.823 \\
\bottomrule
\end{tabular}}
\medskip \\
\resizebox{0.475\textwidth}{!}{
\begin{tabular}{p{2.5cm}ccccc@{}}
\multicolumn{4}{@{}l}{(b) Roof material classification}\\
\toprule
& \textbf{F1 score} & \textbf{Precision} & \textbf{Recall}  & \textbf{Accuracy} \\ \midrule
ConvNeXt-small    & \textbf{0.827} & 0.872  & 0.793 & 0.908  \\
ConvNeXt-base    & 0.819 & 0.830  & 0.809 & 0.904 \\              
ConvNeXt-large    & 0.813 & 0.847  & 0.785 & 0.890 \\
ViT-base    & 0.804 & 0.829  & 0.783 & 0.895 \\              
ViT-large    & 0.802 &  0.845 & 0.772 & 0.894 \\
\bottomrule
\end{tabular}}
\end{table}

These results are consistent with previous findings suggesting that local, country-specific models generally outperform regional models for roof material classification, likely due to variations in roof material distributions across countries, as shown in Table \ref{table:classdist} \cite{tingzon2024mapping}. In contrast, roof pitch appears to be more consistent across countries, allowing additional regional data to improve model performance for roof pitch classification.

\subsection*{Nationwide Structural Builiding Attributes}
Using our best-performing models, we generated country-wide classification maps of roof pitch and roof material for each of the 43,061 building footprints in Saint Vincent and the Grenadines, along with the corresponding probability scores for each prediction. Our results indicate that a majority of the buildings had gable roofs (61\%) and hip roofs (28\%), with 84\% featuring healthy metal roofs, followed by concrete/cement (8\%) and irregular metal (6\%). For the complete statistical breakdown, see Appendix \ref{appendix:b}.

%The dataset is further augmented with additional attributes derived from the building footprint geometries, namely size, length-to-width ratio, and plan irregularity, based on guidelines on the recommended building area and shapes for structural resilience \cite{oecs2025}.

\subsubsection*{Usage and Limitations}
The AI-derived structural buildings attribute dataset is intended as a decision-support tool for aggregated statistical analysis and spatial prioritization. However, it is not a substitute for ground surveys, particularly where high-stakes decisions concerning structural safety or regulatory compliance are involved. This is due to limitations that constrain their applicability for building-level vulnerability analysis. 

One such limitation is the invisibility of critical structural attributes from EO data. Key elements that influence vulnerability, such as roof-to-wall connections and internal structural integrity, limit the model’s reliability for detailed vulnerability assessments. Additionally, the quality of EO inputs affects performance; lower resolution imagery and cloud cover can lead to coverage caps and reduced classification accuracy. Drone-based data acquisition offers an alternative for collecting very high-resolution, cloud-free imagery but depends on local logistical capacity.

We thus emphasize that the AI-derived dataset is not suitable for decision-making at the individual building level without further field verification. Presently, the model outputs are best suited for generating neighborhood-level statistics and supporting the planning of targeted fieldwork. The outputs should therefore be used with caution and an understanding of current technical limitations. 

\section{Conclusion}
This study presents an AI-driven workflow for automated rooftop attribute classification using high-resolution satellite imagery. Our work shows that fine-tuned deep learning models, particularly ConvNeXt variants, outperform ViTs and shallow ML classifiers trained on foundational model embeddings for classifying rooftop attributes in Saint Vincent and the Grenadines. We also demonstrate how incorporating data from neighboring SIDS improved model performance for roof pitch classification but not roof material classification, potentially due to regional variations in roof material distribution. Lastly, we produced nationwide building classification maps and discussed the key limitations of AI-derived structural building attribute datasets. 

%Ultimately, this work provides a scalable framework for SIDS to harness AI and EO data for evidence-based urban planning and disaster risk reduction. Future efforts will focus on integrating multimodal EO data (e.g., street view and drone imagery), improving model robustness to regional variations, and bolstering local capacity to ensure the practical utility of these tools in real-world applications.

%{
%    \small
%    \bibliographystyle{ieeenat_fullname}
%    \bibliography{main}
%}

\appendix
\section*{Appendix}
\addcontentsline{toc}{section}{Appendices}
\renewcommand{\thesubsection}{\Alph{subsection}}

\subsection{Hyperparameter tuning}
\label{appendix:a}
For LR, we implemented grid search CV whereas for SVM and MLP, we used random search CV. For LR, our search space included the norm of the penalty (L1 and L2) and the regularization parameter C (0.001, 0.01, 0.1, 1.0, and 10). For SVM, our search space included the kernel type (linear, polynomial, radial basis function,
and sigmoid), the kernel coefficient gamma (1, 0.1, 0.01, 0.001, and 0.0001), and the regularization
parameter C (0.001, 0.01, 0.1, 1.0, and 10). For MLP, we experimented with different hidden layer sizes, activation functions (tanh and relu), solvers (LBFGS, SGD, and Adam), and regularization parameter alpha (0.0001, 0.001, 0.01, and 0.1). We also experimented with different scaling techniques including standard scaling, min-max scaling, and robust scaling as implemented in scikit-learn \cite{scikit-learn}.

\subsection{Statistics of the building characteristics}
\label{appendix:b}
\begin{figure}[h]
\label{fig:3}
 \centering
 \begin{subfigure}[b]{0.18\textwidth}
     \centering
     \includegraphics[width=\textwidth]{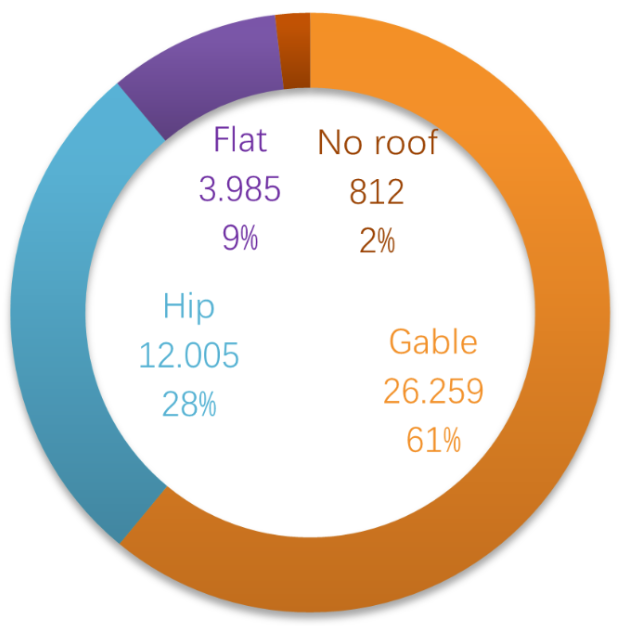}
     \caption{Predicted roof pitch}
 \end{subfigure}
 \hspace{0.75cm}
 \begin{subfigure}[b]{0.20\textwidth}
     \centering
     \includegraphics[width=\textwidth]{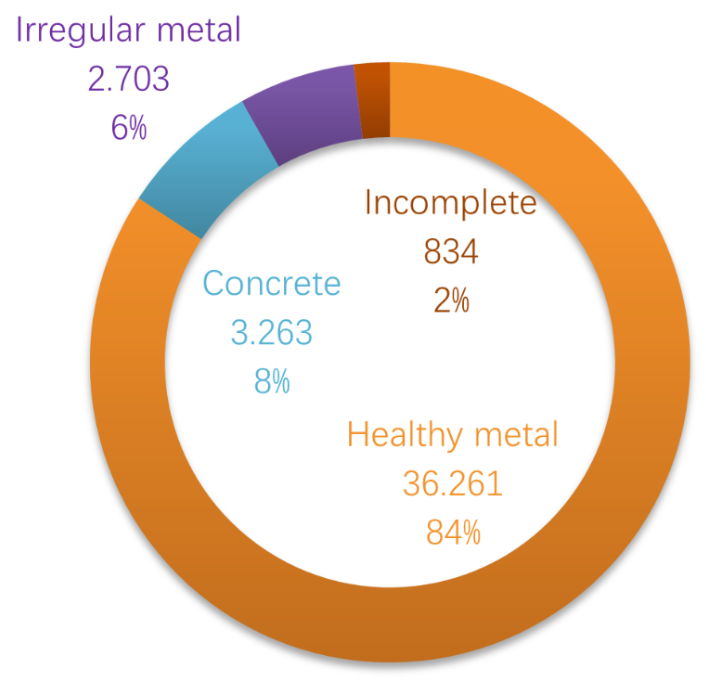}
     \caption{Predicted roof material}
 \end{subfigure}
 \caption{Statistics of the building characteristics in St. Vincent and the Grenadines, indicating predicted roof pitch (top) and predicted roof material (bottom).}
\end{figure}

\end{document}